\pdfoutput=1

\documentclass[11pt]{article}

\usepackage{multirow}
\usepackage{booktabs}
\usepackage{enumitem}

\usepackage{ACL2023} %
\usepackage{graphicx} 

\usepackage{times}
\usepackage{latexsym}
\usepackage[normalem]{ulem}
\usepackage{amsmath,amsfonts,amssymb}
\usepackage{mathtools}

\usepackage[T1]{fontenc}

\usepackage[utf8]{inputenc}

\usepackage{microtype}

\usepackage{inconsolata}

\usepackage{xcolor}  %
\usepackage{soul}
\usepackage{xspace}
\usepackage{epigraph}
\usepackage{tablefootnote}
\usepackage{dsfont}
\usepackage{csquotes}

\usepackage[noabbrev,capitalize]{cleveref}   %

\crefname{page}{page}{pages}
\crefname{footnote}{footnote}{footnotes}   %
\crefname{equation}{equation}{equations}   %
\crefname{corollary}{Corollary}{Corollaries}  %
\crefname{line}{line}{lines}               %
\crefrangeformat{line}{lines #3#1#4--#5#2#6}
\crefname{lstlsting}{Listing}{Listings}   
\crefname{section}{\S}{\S\S}
\Crefname{section}{\S}{\S\S}    %
\crefformat{section}{#2\S#1#3}  %
\Crefformat{section}{#2\S#1#3}
\crefrangeformat{section}{\S\S#3#1#4--#5#2#6}
\Crefrangeformat{section}{\S\S#3#1#4--#5#2#6}
\crefmultiformat{section}{\S#2#1#3}{ and~\S#2#1#3}{, \S#2#1#3}{ and~\S#2#1#3}
\Crefmultiformat{section}{\S#2#1#3}{ and~\S#2#1#3}{, \S#2#1#3}{ and~\S#2#1#3}
\crefrangemultiformat{section}{\S\S#3#1#4--#5#2#6}{ and~\S\S#3#1#4--#5#2#6}{, \S\S#3#1#4--#5#2#6}{ and~\S\S#3#1#4--#5#2#6}
\Crefrangemultiformat{section}{\S\S#3#1#4--#5#2#6}{ and~\S\S#3#1#4--#5#2#6}{, \S\S#3#1#4--#5#2#6}{ and~\S\S#3#1#4--#5#2#6}

\usepackage[]{todonotes}  %
\makeatletter
\newcommand*\iftodonotes{\if@todonotes@disabled\expandafter\@secondoftwo\else\expandafter\@firstoftwo\fi}  %
\makeatother

\newlength{\extramargin}
\usepackage{calc}

\usepackage{subcaption}

\DeclareRobustCommand{\hlcyan}[1]{{\sethlcolor{cyan}\hl{#1}}}
\DeclareRobustCommand{\hlgreen}[1]{{\sethlcolor{green}\hl{#1}}}

\title{Do Androids Know They’re Only Dreaming of Electric Sheep?}

\author{Sky CH-Wang$^{\circ}$\thanks{\ \ Work performed as an intern at Microsoft Semantic Machines. Code and data available at \url{https://github.com/microsoft/llm_generation_probes}.}\quad
Benjamin Van Durme$^{\bullet}$\quad
Jason Eisner$^{\bullet}$\quad
~~Chris Kedzie$^{\bullet}$
\\
$^{\circ}$Department of Computer Science, Columbia University\\
$^{\bullet}$Microsoft\\
\texttt{skywang@cs.columbia.edu, chriskedzie@microsoft.com}
}
\begin{document}
\maketitle

\begin{abstract}

We design probes trained on the internal representations of a transformer language model to predict its hallucinatory behavior on 
three grounded generation tasks. 
To train the probes, we annotate for span-level hallucination on both sampled (organic) and manually edited (synthetic) reference outputs.
Our probes are narrowly trained and we find that they are sensitive to their training domain: they generalize poorly from one task to another or from synthetic to organic hallucinations. However, on in-domain data, they can reliably detect hallucinations at many transformer layers, achieving 95\% of their peak performance as early as layer 4.
Here, probing proves accurate
for evaluating hallucination, outperforming several contemporary baselines and even surpassing an expert human annotator in response-level detection F1. Similarly, on span-level labeling, probes are on par or better than the expert annotator on two out of three generation tasks.
Overall, we find that probing is a feasible and efficient alternative to language model hallucination evaluation when model states are available.

\end{abstract}

\newcommand\blfootnote[1]{%
  \begingroup
  \renewcommand\thefootnote{}\footnote{#1}%
  \addtocounter{footnote}{-1}%
  \endgroup
}

\section{Introduction}\label{sec:intro}

\textit{Do language models know when they're hallucinating?}\footnote{That is, \textit{do they know their electric sheep are ungrounded?}\label{q1label}}
Detecting hallucinations in grounded generation tasks (such as document summarization) is commonly framed as a textual entailment problem
\cite{ji2023survey}, with prior work largely focused on creating or applying secondary detection models trained on and applied to surface text 
(see \citet{falke-etal-2019-ranking,huang-etal-2020-knowledge,kryscinski-etal-2020-evaluating,goyal-durrett-2020-evaluating,mishra-etal-2021-looking}; and others).
In this work, we explore the degree to which the generation model itself---a decoder-only transformer---%
\emph{already} encodes the desired diagnostic information in its feed-forward and self-attention output states.  We train span-level probes supervised at the token-level and response-level probes for hallucination on several grounded generation tasks and show that hallucination with respect to grounding data is detectable from an transformer's hidden states, in line with similar results on probing ungrounded (i.e., closed-book QA) generation \cite{mielke-etal-2022-reducing,azaria2023internal}. 

\begin{figure}[t]
    \centering
    \includegraphics[width=0.32\textwidth]{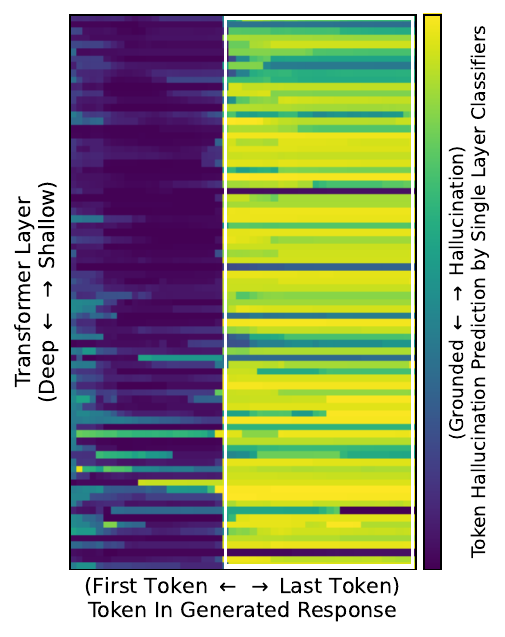}
    \caption{Prediction of hallucination in one generated response by probing the hidden states of a transformer during decoding. The true annotated span of hallucination within the response is boxed in white; rows represent probes trained to detect hallucination from different transformer layers, with columns representing their prediction for each token in a generated response.}
    \label{fig:example_tokenclassification}
\end{figure}

As a language model generates a response, we apply our probes to its encoded tokens to determine whether the current token
is likely to be a hallucination. Related work has argued that these encodings are full of meta-information about the model's generation process. For example, salient information for generation can be localized to specific model layers and time steps \cite{geva-etal-2021-transformer,meng2022locating, meng2023mass}, and hidden states can encode the model's confidence about the correctness of subsequent response generation \cite{mielke-etal-2022-reducing}. This suggests that a probe on model hidden states will have access to a rich set of features for detecting hallucination—features that would not be directly and efficiently available to a secondary entailment or classification model seeing only the input and generated output tokens.

We develop three probes of increasing complexity across three grounded generation tasks: abstractive summarization, knowledge-grounded dialogue generation, and data-to-text.
For each task, we collect hallucinations in two ways:
 (1)~from ordinary sampled responses, where we generate outputs from a large language model (LLM) conditioned on the inputs, or (2)~by editing reference inputs or outputs to create discrepancies.  
We refer to these as organic and synthetic data, respectively.  In both cases, we have human annotators mark hallucinations in the outputs.
Method (2) produces hallucination annotations at a higher rate, though we find that the utility of these synthetic examples is lower as they do not come from the test distribution.

We summarize our contributions as follows:

\begin{itemize}
\setlength\itemsep{0em}
\item We produce a high-quality dataset of more than 15k utterances with hallucination annotations for organic and synthetic output texts across three grounded generation tasks.

\item We propose three probe architectures for detecting hallucinations and demonstrate improvements over multiple contemporary baselines while also surpassing an expert human annotator on response-level detection F1. On two out of three tasks, the best probe achieves over 90\% F1. 
\item We show probe performance on span-level detection is competitive with an expert human annotator, performing on par or better on two out of three generation tasks in our dataset.
\item We analyze how probe accuracy is affected by annotation type (synthetic/organic),
hallucination type (extrinsic/intrinsic),
model size, and which part of the encoding is probed.
\end{itemize}

\section{Related Work}
\label{relatedwork}

\paragraph{Definitions.} 
We center our study of hallucinations in the setting of in-context generation \cite{lewis2020retrieval} where
grounding knowledge sources are provided within the prompt itself. This is in contrast to settings where a language model is expected to generate a response entirely from its parametric
knowledge (i.e., grounded in the model's training data, as in \citet{azaria2023internal}).
Though the term has often appeared in reference to factuality \cite{gabriel-etal-2021-go}, hallucinations---as we use the term---are not necessarily factually incorrect, only ungrounded. 
We follow the taxonomy of \citet{ji2023survey}: in their survey, they classify hallucinations as either \textit{intrinsic}, where generated responses directly contradict the knowledge sources, or \textit{extrinsic}, where generated responses are neither entailed nor contradicted by the sources.
We follow this taxonomy.
For example, given the following task prompt for abstractive summarization,
\begin{displayquote}
    \dots \small{Deckard is now able to buy his wife Iran an authentic Nubian goat—from Animal Row, in Los Angeles—with his commission} \dots \small{TL;DR:} %
\end{displayquote}
a generated summary sentence of
``\textit{Deckard can now purchase an animal for his wife}''
is faithful,
``\textit{Deckard can purchase a goat in Iran}''
is an intrinsic hallucination, and
``\textit{A Nubian goat is a British domestic goat breed}''
is an extrinsic hallucination. 
While models may leverage latent knowledge to \textit{inform} their responses,
such as recognizing that a Nubian goat is indeed a British domestic goat breed,
directly \textit{stating} such knowledge in a response—as in  
the last sentence above—is
considered an extrinsic hallucination when it is neither entailed nor contradicted by the knowledge sources. Thus, non-hallucinatory generation in this setting involves balancing the language model's retrieval and manipulation of knowledge, both from the given prompt context and from its parameters.

\begin{table*}
\centering
    \begin{tabular}{ p{1cm} p{2.1cm} p{4.2cm} p{1.8cm} p{4.7cm} }
        \toprule
        \small{\textbf{Task}} & \small{\textbf{History}} & \small{\textbf{Knowledge}} & \small{\textbf{Instruction}} & \small{\textbf{Response}} \\ 
        \midrule
        \small{\textbf{Abs. Summ.}} & -- &                           \small{...dialed down that initial forcast to predict 13 to 19 named storms, including six to nine hurricanes...} &   \small{TL;DR:} &                \small{The National Weather Service forecast called for 13 to 19 named storms including six to \hlcyan{11} hurricanes.} \\
        \midrule
        \small{\textbf{KGDG}} & \small{...B: ok how can I get a turkey I think that may be hard to get though} &     \small{Three consecutive strikes are known as a ``turkey.''} &     \small{Response that uses the information in knowledge:} &    \small{A: Indeed, you have to get \hlcyan{four} consecutive strikes to get one.}  \\
        \midrule
        \small{\textbf{D2T}} & &                           \small{name[The\,Wrestlers], priceRange[high]} &          \small{Restaurant Description:} &           \small{\hlgreen{Located by the riverside,} The Wrestlers is a restaurant with high prices, \hlgreen{and near Raja Indian Cuisine.}} \\ 
        \bottomrule
    \end{tabular}
\caption{From top to bottom, examples of organic hallucinations in model responses for \textbf{abstractive summarization}, \textbf{knowledge-grounded dialogue generation} (KGDG), and \textbf{data-to-text} tasks. Minimum annotated hallucination spans are highlighted in \hlcyan{cyan} for intrinsic hallucinations and \hlgreen{green} for extrinsic hallucinations.}
\label{tab:example_hallucination}
\end{table*}

\paragraph{Automated Diagnosis.}
Lexical metrics commonly used in the evaluation of NLG fail to correlate with human judgments of faithfulness \cite{kryscinski-etal-2019-neural}.  More useful are natural language inference (NLI) approaches that either directly measure the degree of entailment between a generated response and a given source text \cite{hanselowski-etal-2018-ukp, kryscinski-etal-2020-evaluating, goyal-durrett-2020-evaluating}, or else break down texts into single-sentence claims that are easily verifiable \cite{factscore, semnani2023wikichat}. These NLI methods have been successfully used for hallucination detection \cite{laban-etal-2022-summac, bishop2023longdocfactscore}. Similarly, approaches that leverage question-answer (QA) models to test if claims made by 
a response can be supported
by the grounding text have also been applied to measure hallucination \cite{wang-etal-2020-asking, durmus-etal-2020-feqa, scialom-etal-2021-questeval}.
To the extent that models are less confident when they are hallucinating, hallucination may also be reflected in metrics that measure generation uncertainty \cite{malinin2021uncertainty, guerreiro-etal-2023-looking, ren2023outofdistribution, ngu2023diversity, madsen2023faithfulness}, relative token contributions to the overall output probability \cite{xu-etal-2023-understanding}, or 
signatures in the attention matrices
during generation \cite{lee2018hallucinations,attnsats}.
Similarly, conflicts between contextual knowledge provided in a prompt and parametric knowledge as learned during pre-training can contribute to this behavior \cite{xu2024knowledge}.

\paragraph{Predicting Transformer Behavior.}
In small transformer language models, it is possible to ascribe behaviors like in-context learning to specific layers and modules \cite{elhage2021mathematical}. 
 This becomes harder in larger models, but \citet{meng2022locating}
 provide evidence that the feed-forward modules function as key-value stores of factual associations and that it is possible to modify weights in these modules to edit facts associated with entities. 
Transformer internal states 
have also been shown to be predictive of the likelihood that a model will answer a question correctly in the setting of closed-book QA \cite{mielke-etal-2022-reducing} and in the prediction of the truthfulness of a statement \cite{azaria2023internal}. Interestingly, these internal representations of truthfulness can disagree systematically with measures of model confidence \cite{liu-etal-2023-cognitive}. 
These works
measure hidden states in the absence of supporting evidence (i.e. no passages are retrieved for either task), making them effectively probes of knowledge stored in model weights and not the manipulation of novel or contingent information provided in a prompt. In contrast, this work examines related questions specifically in the context of grounded generation (e.g., retrieval augmented generation \cite{lewis2020retrieval}).

Concurrently to this work, \Citet{monea2023glitch} 
find that there are separate computational processes in transformer LLMs devoted to the recall of associations from memory versus grounding data.
\Citet{slobodkin-etal-2023-curious} train a probe to detect the (un)answerability of a question given a passage, which is similar in some respects to our task in \cref{probedesign}; however, we detect span-level hallucination during answer generation, which could happen even in theoretically answerable questions.\footnote{Reviewers asked to distinguish our work from that of \citet{azaria2023internal} and \citet{kadavath2022language}. For a detailed discussion of the differences, see \cref{sec:expandedresults}.}

\section{Grounded Generation Tasks}\label{sec:tasks}

\begin{table*}
\centering
\begin{tabular}{ l l l }
    \toprule
    \textbf{Response} & \textbf{Original Knowledge} & \textbf{Changed Knowledge} \\ 
    \midrule
    Yes, Elvis was also one of      & Elvis Presley is regarded as one of           & Elvis Presley is regarded as one \\
    the \hlcyan{most famous musicians}       & the \textbf{most significant cultural icons}  & of the \textbf{least well-known music-}  \\ 
    of the 20th century.            & of the 20th century.                          & \textbf{-ians} of the 20th century. \\
    \midrule
    Close to The Sorrento you       & name[Clowns], \textbf{eatType[pub]}           & name[Clowns], \\
    can find Clowns \hlgreen{pub}.            & near[The Sorrento]                            & near[The Sorrento] \\
    \bottomrule
\end{tabular}
\caption{Examples of changed knowledge statements for our datasets of synthetic hallucinations. Changes made to Conv-FEVER (top) and E2E (bottom) are shown in \textbf{bold}. The changed \textit{knowledge} statements make the top \textit{response} an \hlcyan{intrinsic} predicate error and the bottom \textit{response} an \hlgreen{extrinsic} entity error (highlighted as in \cref{tab:example_hallucination}).}
\label{tab:synthetic}
\end{table*}

We test hallucination probes for autoregressive grounded generation in three distinct tasks:~ abstractive summarization, knowledge-grounded dialogue generation, and data-to-text. 
We use the same high-level prompt structure across all tasks to organically generate model responses to tasks and to probe for hallucination. Each prompt contains
an optional interaction \textit{history}, a \textit{knowledge} source, and an \textit{instruction} resulting in the conditionally generated \textit{response}. Task prompts and examples of organic hallucinations
are shown in \cref{tab:example_hallucination}.

\subsection{Organic Hallucinations}

We use \texttt{llama-2-13b} \cite{touvron2023llama} as the response model
to generate outputs for each task. To ensure that a non-negligible number of both hallucinations and grounded responses were present in the generations, each model response is generated using top-\textit{k} random sampling \cite{fan-etal-2018-hierarchical} with $k=2$ and temperature $1$.
The number of few-shot in-context learning examples for each task (see below) was chosen through a pilot exploration, optimizing for the presence of a balance of hallucinations and fully grounded responses, as determined via manual inspection by the authors.
Full prompt templates are shown in \cref{sec:prompts}.

\paragraph{Abstractive Summarization.}
The CNN / DailyMail (CDM) dataset  \cite{hermann2015teaching} is a standard benchmark in the setting of abstractive summarization \cite{lin2019abstractive}. It consists of a set of news articles paired with reference summaries in the form of human-written bullet points that were originally published alongside the articles. For automatic summarization, we follow \citet{radford2019language} and prompt our model to generate summaries of news articles by appending \texttt{TL;DR:} after the article content (a 0-shot prompt),\footnote{In Internet posts that appear in LLM pretraining data, this string often marks the start of a summary immediately following the main content of the post. Prior work \cite{radford2019language} has found that appending this string after texts in prompts to autoregressive LLMs is capable of producing abstractive summaries of the former.} taking the first three generated sentences as the summary.

\paragraph{Knowledge-Grounded Dialogue Generation.}
For a given conversation history and a provided knowledge sentence obtained via Wizard-Of-Wikipedia, the Conv-FEVER \cite{santhanam2021rome} task (CF) asks a respondent to create a next-turn response to the conversation that is grounded in
the provided knowledge sentence.
Here, we prompt our response model to generate grounded responses under 2-shot in-context learning with example grounded responses.

\paragraph{Data-to-Text Generation.} The E2E task \cite{novikova-etal-2017-e2e} asks respondents to write restaurant descriptions that are congruent with a provided set of attributes. Here, we prompt our response model to 
perform this task with 9-shot in-context learning, with in-context examples manually validated for groundedness.\footnote{As the original E2E task had omitted or missing information in up to 40\% of its data, we use the cleaned version by \citet{dusek-etal-2019-semantic}. As even the cleaned version had instances of omitted or missing information, 9 in-context examples were manually selected and verified for groundedness.}

\subsection{Synthetic Hallucinations}

Recent work \cite{ma-etal-2023-bump, monea2023glitch} 
has carefully created synthetic ungrounded
datasets by hand to evaluate metrics used to measure unfaithfulness and models used to detect it.
The BUMP dataset \cite{ma-etal-2023-bump} is a version of CDM in which reference summaries have been manually edited to be ungrounded.
It includes both freestyle edits and systematic coverage of four specific types of errors: (1) \textit{predicate errors}, where the predicate in a generation is inconsistent with the source knowledge; (2) \textit{entity errors}, where the subject or object of a predicate is inconsistent; (3) \textit{circumstance errors}, or errors in the time, duration, or location of an event of the predicate; and (4) \textit{coreference errors}. Types (1)--(3) may be intrinsic or extrinsic, while (4) may be ambiguous; as such, we don't consider coreference errors in our analysis of intrinsic or extrinsic hallucinations.

We use BUMP in our CDM experiments. For our Conv-FEVER and E2E experiments, we similarly created errors in a uniformly sampled subset of these datasets.
However, rather than changing an example's reference response,
we changed its provided knowledge sentence so that the existing reference response would be unfaithful to it. This simulates the common case of hallucination where a large language model generates statements that may be 
plausible but are not technically grounded.

For Conv-FEVER, we prompted ChatGPT to modify a given example's provided knowledge sentence to introduce intrinsic and extrinsic errors of types (1)--(4) above;
we then manually judged these attempts and edited the unsuccessful ones.\footnote{In hindsight, this workflow proved inefficient; significant human editing and validation were needed.}
In this way, we created equal numbers of examples with intrinsic and extrinsic predicate, entity, circumstance, and coreference errors, as well as freestyle errors.

For E2E, we modified a random subset of the provided grounding attributes. Specifically, for an example with $n$ attributes, we drew $k$ uniformly from $[1,n-1]$, uniformly selected one of the subsets of size $k$, and then flipped a fair coin to decide whether to remove or perturb—by randomly sampling an alternative attribute from the entire dataset—these $k$ attributes.
Finally, we manually verified that the reference response was unfaithful with respect to the modified attributes; if not, we further edited the attributes.
Examples of changed knowledge statements are shown in \cref{tab:synthetic};
statistics of our synthetic datasets are shown in \cref{tab:annotation_stats}.

\begin{table}
\centering
\begin{tabular}{ l  l l l l l }
    \toprule
    \textbf{Task} & \textbf{N-o} & \textbf{H-o} & \textbf{$\kappa$-o}  & \textbf{N-s} & \textbf{H-s} \\ 
    \midrule
    CDM &   4684 &      28\% & 0.39 &           1778\tablefootnote{Obtained via BUMP \cite{ma-etal-2023-bump}.} & 50\% \\  
    CF &     924 &      70\% & 0.50 &           2846 & 50\% \\
    E2E &   4998 &      54\% & 0.55 &           9950 & 50\% \\
    \bottomrule
\end{tabular}
\caption{\textbf{Dataset statistics.} \textbf{N-o} is the number of annotated organic generations, \textbf{H-o} is the percentage of organic generations with at least one hallucination in the final adjudicated dataset, and \textbf{$\kappa$-o} is the response-level Fleiss's $\kappa$ of individual annotations before adjudication. \textbf{N-s} and \textbf{H-s} refer to the same for synthetic responses. Note that grounded synthetic responses were not generated by a model but come from gold reference data.\looseness=-1}
\label{tab:annotation_stats}
\end{table}

\subsection{Annotation}\label{sec:annotation}
 
We ask annotators to label the \textit{minimal hallucinatory spans} in each response, if any.  These are the smallest text spans that would need modification to transform an ungrounded response into a grounded one.
A team of 17 annotators completed a pilot annotation exercise of 50 annotations per task.  These pilot annotations were checked and verified by the authors to pinpoint common sources of error and disagreement. This feedback was integrated into a revised set of annotation guidelines, shown in \cref{sec:annotationguidelines}, and the full dataset was then re-annotated fully by three annotators such that each example was annotated three times
(example annotations are shown in \cref{tab:example_hallucination}).
Annotations were then reconciled at the  token-level into the final gold annotation, such that a token is labeled as a hallucination if at least two out of three annotators included it in a hallucination-span annotation. 

Annotation statistics are shown in \cref{tab:annotation_stats}.
Note that for each dataset, over one-fifth of all annotated responses were annotated as containing at least one hallucinated span (columns \textbf{H-o} and \textbf{H-s}), which we refer to as response-level annotations. Annotators exhibited moderate agreement \cite{landis1977measurement}, achieving response-level Fleiss' $\kappa$ scores between 0.39 and 0.55 across the three tasks.\looseness=-1

\section{Probing}
\label{probedesign}

Probes \cite{alain2017understanding} are tools used to analyze a neural network's internal representations. Often taking the form of linear classifiers, they are applied to the internal representations of a network and are trained to discriminate between different types of inputs or outputs. Here, we are interested in detecting hallucinatory outputs during an LLM's generation steps when conditioned on a prompt that requires a response grounded in the prompt.

\subsection{Design} 

Let $u$ be a prompt as described at the start of \cref{sec:tasks}. To determine whether a response $x$ to a prompt $u$ contains hallucinations, we will probe the hidden states of the transformer language model (i.e., decoder) as it generates $x$. Note that these hidden states depend on $u$ and on the prefix of $x$ that has been generated so far, and they may be reconstructed from the string pair $(u,x)$ via forced decoding.
A single transformer layer actually consists of two sublayers \cite{vaswani2017attention}: the first transforms a token's embedding by adding a vector predicted by attention, and the second transforms it again by adding a vector predicted by a feed-forward network.  
We train a separate probe for each of the $2L$ sublayers, where $L$ is the number of layers.
\looseness=-1

\paragraph{Linear Probe.} This probe is a linear classifier
using a single hidden state vector (output by some sublayer) as input. Let $x_i$ be the $i^\textrm{th}$ response token and let   $\mathbf{h}_i \in \mathbb{R}^n$ be the corresponding LLM hidden state at a fixed layer and component.
We write $y_i \in \{0,1\}$ to indicate whether  $x_i$ falls within a minimum hallucination span.  
The linear probe predicts $y_i$ by
 $p(y_i=1\mid u, x_{\leq i}) = \sigma(\mathbf{w}^\intercal \mathbf{h}_i + b)$, where $\mathbf{w} \in \mathbb{R}^n$ are learned weights, $b \in \mathbb{R}$ is a learned bias, and $\sigma$ is 
 the logistic function.

\paragraph{Attention-Pooling Probe.} In order to directly consider the previous hidden states
of the response as well,
we also experiment with an attention-pooling probe over all hidden states so far,
\[p(y_i=1\mid u, x_{\leq i}) = \sigma(\mathbf{w}^\intercal \mathbf{\bar{h}}_i ) \]
\[\mathbf{\bar{h}}_i = \sum_{j=1}^i \alpha_{i,j} \mathbf{h}_j, 
\;\; \alpha_{i,j} = \frac{\exp\left(\mathbf{q}^\intercal \mathbf{h}_j\right)}{\sum_{k=1}^i \exp\left( \mathbf{q}^\intercal \mathbf{h}_k\right)   }  \]
where $\mathbf{q} \in \mathbb{R}^n$ is a learned query vector.
\paragraph{Ensemble Probe.} Our final approach is a %
logistic regression probe,
$p_{\textrm{ens}}(y_i\mid u,x_{\leq i}) = \sigma\left( \sum_{l,m}\beta_{l,m} \cdot p_{l,m}(y_i\mid u,x_{\leq i}) \right)$,
that combines the predictions from $2L$ separately trained probes 
indexed by layer $l \in \{1,\ldots,L\}$ and module $m \in \{\textrm{feed-forward}, \textrm{attention}\}$.
The weights $\beta_{l,m}$ are learned after training and freezing each $p_{l,m}$.

\subsection{Training Objectives}

\paragraph{Token-level.} To train a probe to predict \emph{token-level} hallucination labels $y_i$, we fit probe parameters to minimize
the negative log-likelihood,
$\mathcal{L}(\theta) \propto \sum_{(u,x,y) \in \mathcal{D}}\sum_{i=1}^{|x|} -\log p(y_i\mid u, x_{\leq i}; \theta),$
on the token-level annotations in our dataset $\mathcal{D}$.

\paragraph{Response-level.}  We also try training a classifier to predict whether $x$ contains any hallucinations at all;   
denote this indicator as $y = \bigvee_{i=1}^{|x|} y_i \in \{0,1\}$.  
For this, we simply use an attention-pooling probe over the entire response $x$.  To fit the 
probe parameters, $\theta$, we again minimize negative log-likelihood,
$ \mathcal{L}(\theta) \propto \sum_{(u,x,y) \in \mathcal{D}} -\log p(y\mid u,x_{\leq |x|}; \theta).$

\subsection{Metrics}\label{sec:metrics}

\paragraph{Response-level Classification.} F1 is the harmonic mean of recall and precision.  We denote the F1 score of a response-level classifier by \textbf{F1-R}.

\paragraph{Span-level Classification.}
For span-level classifiers, it is possible to evaluate the F1 score of their per-token predictions; however, this would give more weight to annotated or predicted spans that are longer in length. Here, we are more interested in \textit{span-level} F1: specifically, whether we predicted all of the annotated hallucinations (recall), and whether all of the predicted spans were annotated as actual hallucinations (precision). As requiring an exact match of  spans is too severe—\cref{sec:annotatordisagreements} shows how our human annotators often disagreed on the exact boundaries of a hallucinatory span—we modify span-level precision and recall metrics to give partial credit, as follows.

Define a span to be a set of tokens, where tokens from different responses are considered distinct. 
Let $A$ be the set of all annotated spans in the reference corpus, and let $P$ be the set of all predicted spans on the same set of examples.  
We compute the recall $r$ as the average coverage of each reference span, i.e., the fraction of its tokens that are in some predicted span.  Conversely, we compute the precision $p$ as the average coverage of each predicted span. So
$r = \frac{1}{|A|} \sum_{s \in A} \frac{ 
 \left| s \cap \left(\bigcup_{\hat{s} \in P} \hat{s} \right) \right|  
}{|s|}$ and 
$p = \frac{1}{|P|}\sum_{\hat{s} \in P} \frac{ \left| \hat{s} \cap \left(\bigcup_{s \in A} s \right) \right| }{|\hat{s}|}$
where $\cup$, $\cap$, and $|\cdot|$ denote set union, intersection, and cardinality respectively.
We define \textbf{F1-S$_p$} (standing for "span, partial credit") as the harmonic mean of $p$ and $r$.
Note that these quantities micro-average over spans, ignoring the boundaries between examples. Thus, in contrast to \textbf{F1-R}, examples with more spans have more influence on the final \textbf{F1-S$_p$} metric achieved.

\section{Experiments}

Unless otherwise indicated, all evaluations are performed on organic data only, even if their probes were trained on synthetic data.

\subsection{Probe Hyperparameters and Training}

Single-layer hidden state probes, described in \cref{probedesign}, are trained under \texttt{float32} precision with unregularized log loss. Optimal batch sizes between 20-100 
and Adam optimization \cite{kingma2014adam}—with learning rates between 0.1-0.001—and $\beta_1, \beta_2 = (0.9, 0.999)$ were chosen for each task after a small grid search on hyperparameters. With 70/10/20\% train, validation, and test splits, each classifier was trained using early stopping with 10-epoch patience. 
The ensemble classifier was trained with the same hyperparameters as the single-layer probes. 

For CDM, the synthetic training dataset (from BUMP) is
smaller than the organic training dataset, so the SYNTH results
are at a disadvantage. For CF and E2E, however, we generated
a synthetic training dataset that was matched in size.

When reporting results, 
\textit{SL} denotes the best single-layer classifier (chosen on the validation set), \textit{E} denotes the ensembled classifier, and \textit{SYNTH} indicates that the classifier was trained on synthetic data and tested on organic hallucinations.

\subsection{Response-Level Baselines}

We compare the performance of our probes against a set of contemporary methods
used in faithfulness evaluation, measuring test set hallucination classification F1.
Specifically: 
(1)~the model uncertainty-based metric of length-normalized sequence log-probability \textbf{Seq-Logprob} \cite{guerreiro-etal-2023-looking};
(2)~%
off-the-shelf synthetically-trained sentence-level claim classification \textbf{FactCC} \cite{kryscinski-etal-2020-evaluating};
(3)~linguistic feature-based dependency arc entailment \textbf{DAE} \cite{goyal-durrett-2020-evaluating};
(4)~QA-based metrics \textbf{FEQA} \cite{durmus-etal-2020-feqa}; (5)~\textbf{QuestEval} \cite{scialom-etal-2021-questeval};
(6)~the NLI-based metric \textbf{SummaC} \cite{laban-etal-2022-summac};
\textbf{ChatGPT} classifiers \cite{luo2023chatgpt} \texttt{gpt-3.5-turbo-0125} at zero temperature (7) without and (8) with basic chain-of-thought prompting (with prompts shown in \cref{sec:chatgptprompts}), as well as (9) the automatic claim-breaking and verification method \textbf{FActScore} \cite{factscore} with \texttt{gpt-3.5-turbo-instruct}. 

Metrics (2)--(6) and (9)
were not originally intended to evaluate individual responses, but rather a collection of responses for the sake of system comparisons.  To adapt to our setting, however, we threshold each of these continuous metrics to classify the individual responses, setting the threshold to maximize validation set F1, following \citet{laban-etal-2022-summac}. 
For sentence-level optimized baselines such as FactCC, response-level evaluation is performed by logical \texttt{OR}-ing sentence-level baseline predictions (obtained after thresholding the raw scores). In other words, any hallucination in any sentence leads to a "hallucinated" label.
Note that, for our abstractive summarization tasks, methods such as FactCC \cite{kryscinski-etal-2020-evaluating} are explicitly trained in the domain as they use CDM as the initial dataset for training. For certain baselines in CF and E2E, dashes are present in the results due to an inability to directly compare methods in these domains. For example, for DAE, one cannot obtain the dependency arcs from a set of restaurant attributes not expressed in natural language. 

As a further reference baseline, we additionally report the performance achieved by randomly predicting a hallucination label with some probability \textit{p} tuned to maximize F1 on the validation set,
denoted here as Optimized Coin (\textbf{OC}).

\subsection{Expert Human Judge}
To compare our probe performance to a human performing the equivalent hallucination detection task, we had a another professional annotator judge the test set. This annotator (denoted \textbf{Expert Human}) was 
distinct from the annotators who produced the gold labels. They were given access to the training data (consisting of all three sets of annotations plus the gold reconciled annotations) and performed a round of pilot annotations with feedback from the authors before annotating the test set.

\subsection{Results}

\begin{table}
\centering
\begin{tabular}{ l  c c c }
    \toprule
    \multicolumn{4}{c}{\textbf{Response-Level Classification, F1-R}} \\
    \midrule
    Method & CDM & CF & E2E \\
    \midrule
    OC & 0.43 & 0.79 & 0.71 \\
    Seq-Logprob & 0.64 & 0.81 & 0.72 \\
    FactCC & 0.51 & - & - \\
    DAE & 0.53 & - & - \\
    FEQA & 0.44 & - & - \\
    QuestEval & 0.61 & - & - \\
    SummaC$_{ZS}$ & 0.68 & - & - \\
    SummaC$_{Conv}$ & 0.67 & - & - \\
    ChatGPT & 0.63 & 0.74 & 0.74 \\
    ChatGPT CoT & 0.53 & 0.51 & 0.47 \\
    FActScore & 0.61 & 0.80 & 0.72 \\
    \midrule
    Pooling\textsubscript{SL} & 0.73 & 0.93 & 0.89 \\
    Pooling\textsubscript{E} & \textbf{0.75} & \textbf{0.94} & \textbf{0.90} \\
    Pooling\textsubscript{SL-SYNTH} & 0.65 & 0.80 & 0.68 \\
    Pooling\textsubscript{E-SYNTH} & 0.66 & 0.83 & 0.71 \\
    \midrule
    Expert Human & 0.55 & 0.89 & 0.79 \\
    \bottomrule
\end{tabular}
\caption{\textbf{Response-level} probe and baseline organic hallucination detection performance across tasks as measured via F1 scores (\textbf{F1-R}) achieved on the test set. 
}
\label{tab:mainresults_response}
\end{table}

\looseness=-1
For performance comparisons, probe results are statistically significant ($p<0.05$) relative to respective baselines under exact paired-permutation testing \cite{zmigrod-etal-2022-exact} unless otherwise noted.

\looseness=-1
\paragraph{Response-level Results.}
Response-level results are shown in \cref{tab:mainresults_response}. On all three tasks, the probes were able to surpass the Expert Human annotator on response-level classification, with the Pooling\textsubscript{E} probe outperforming the Expert Human by as much as 20 points absolute F1-R on
the CDM dataset. The CDM dataset is likely the most challenging for a human
to judge as it requires checking the generated statements against 
the complete (and lengthy) news article.
Both Pooling\textsubscript{SL} and Pooling\textsubscript{E} probes outperformed all baselines across all datasets. Ensembling generally adds at least one point in absolute F1-R above the best single layer. %
\looseness=-1
The probes trained on synthetic hallucinations generally do not perform as well as their organic equivalents, though they are still fairly competitive with the baselines. 
Seq-Logprob was surprisingly strong, beating both the Expert Human and all but one of the baselines on CDM. This suggests that while model confidence may not be the only feature related to hallucination detection, it is likely an important one. 

\looseness=-1
\paragraph{Span-Level Results.} These results appear in \cref{tab:mainresults_token}.
The Pooling\textsubscript{E} probes match or surpass the Expert Human 
performance on the CDM and E2E datasets while falling three points behind in absolute F1-S\textsubscript{p} on the Conv-FEVER dataset. Pooling information across time does help probe performance with the best single-layer pooling probe matching or beating the best single-layer linear probe across all datasets. Interestingly, what the linear probe lacks in time can be made up in depth---Linear\textsubscript{E} matches Pooling\textsubscript{SL} on CDM and Conv-FEVER, and beats it by one point absolute F1-S\textsubscript{p} on E2E, though this wasn't statistically significant.
\looseness=-1

\section{Analysis}

\begin{figure}[t]
    \centering
    \includegraphics[width=0.48\textwidth]{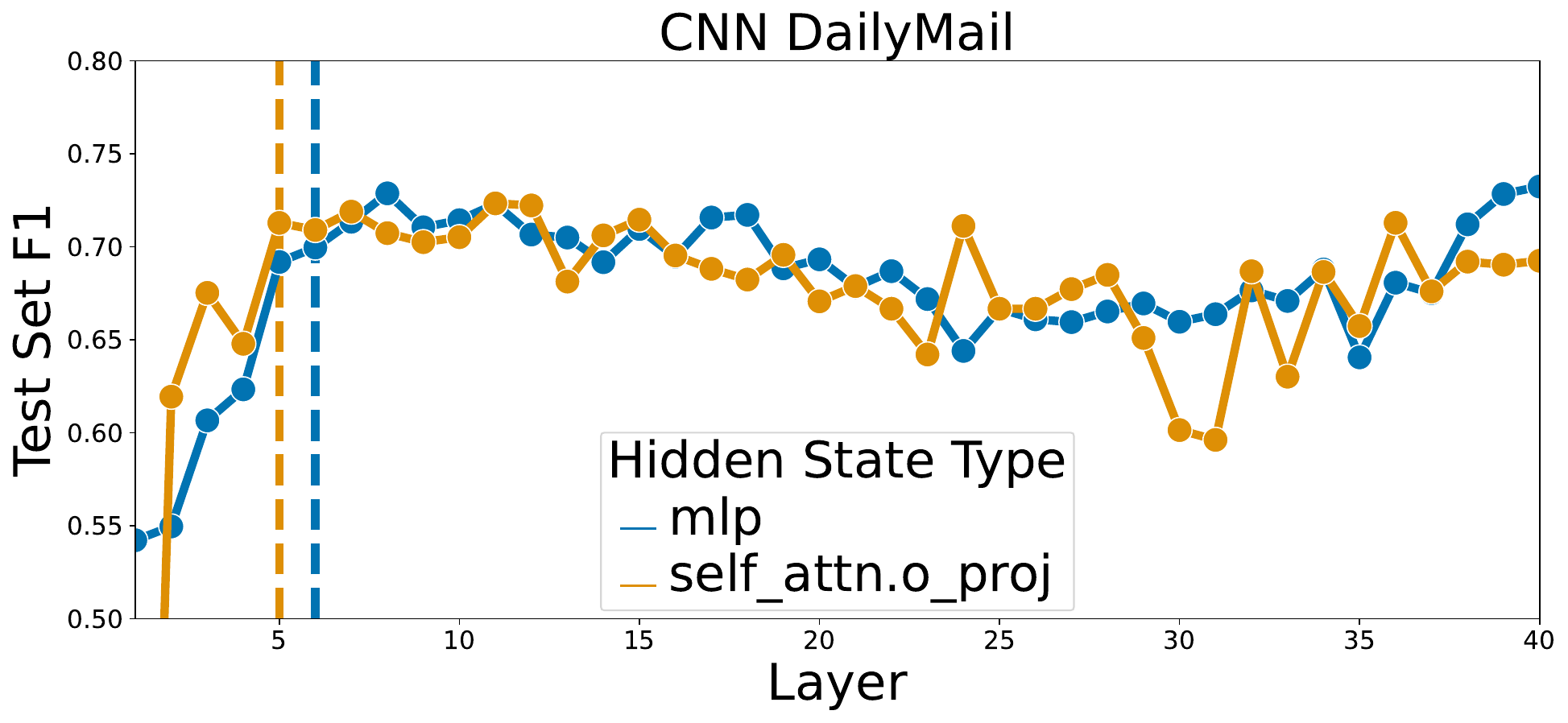}
    \caption{Grounding behavior saliency as measured by probe response-level F1 (\textbf{F1-R}) on the CDM test set. The $x$-axis corresponds to the layer of the probe. Plots across all tasks are shown in \cref{sec:expandedresults,fig:layers_types_all}. Vertical dashed lines denote the layers that respective probes first surpass 95\% of their peak performance.
    }
    \label{fig:layers_types}
\end{figure}

\paragraph{Layers.}
Evaluating the performance of trained single-layer attention-pooling probes (\cref{fig:layers_types,fig:layers_types_all}),
in model \textit{middle layers} during decoding. Probe performance shows a consistent and gradual increase from the first layer after token embeddings
up until layers 10 to 20. 
Hallucination information is reliably present at all layers $\geq 5$, although there are local optima around layers 5--10 and again at 40.  Ensembling across many layers provides only a small (not stat. sig.) improvement over the best 
single layer (\cref{tab:mainresults_response}), so we do not have evidence that our probes at different layers detect different types or correlates of hallucination.
In general, probe performance declines for subsequent layers after their initial peak layer across tasks, though CDM is an exception in that probe performance gradually increases again from layers 30 to 40, peaking there.

\paragraph{Hidden State Type.} Slight variations in grounding saliency—the strength of the signal of hallucination as able to be detected by a probe, measured by probe hallucination classification performance—are also present among hidden states of different types during decoding. As shown in \cref{fig:layers_types}, feed-forward probes reach peak performance at slightly deeper layers than their attention head out-projection probe counterparts and exceed them in single-layer hallucination detection. When training ensemble classifiers on only the hidden states of the same type at every layer, feed-forward hidden states possess a slightly greater \& statistically significant saliency of model grounding behavior.

\begin{table}
\centering
\begin{tabular}{ l c c c }
    \toprule
    \multicolumn{4}{c}{\textbf{Span-level Classification, F1-S$_p$}} \\
    \midrule
    Method & CDM & CF & E2E \\
    \midrule
    OC & 0.31 & 0.75 & 0.13 \\
    Linear\textsubscript{SL} & 0.52 & 0.77 & 0.54 \\
    Linear\textsubscript{E} & 0.54 & 0.79 & 0.55 \\
    Pooling\textsubscript{SL} & 0.54 & 0.79 & 0.54 \\
    Pooling\textsubscript{E} & \textbf{0.55} & \textbf{0.81} & \textbf{0.56} \\
    \midrule
    Expert Human & 0.40 & 0.84 & 0.56 \\
    \bottomrule
\end{tabular}
\caption{\textbf{Span-level} probe performance, as measured by span-level partial match scores (\textbf{F1-S$_p$}). 
}
\label{tab:mainresults_token}
\end{table}

\paragraph{Model Size.} Training probes
in the same manner on the force-decoded hidden states of responses on our dataset using \texttt{llama-2-7b} and \texttt{llama-2-70b}, we find no statistically significant
difference 
in hallucination detection
across the hidden states from base models of varying sizes. %
These results show that we can still detect a hallucination even when it was originally generated from a different model, and that even the lower-dimensional hidden states of the 7B model still contain enough information to train a probe.

\begin{figure}[t]
    \centering
    \includegraphics[width=0.48\textwidth]{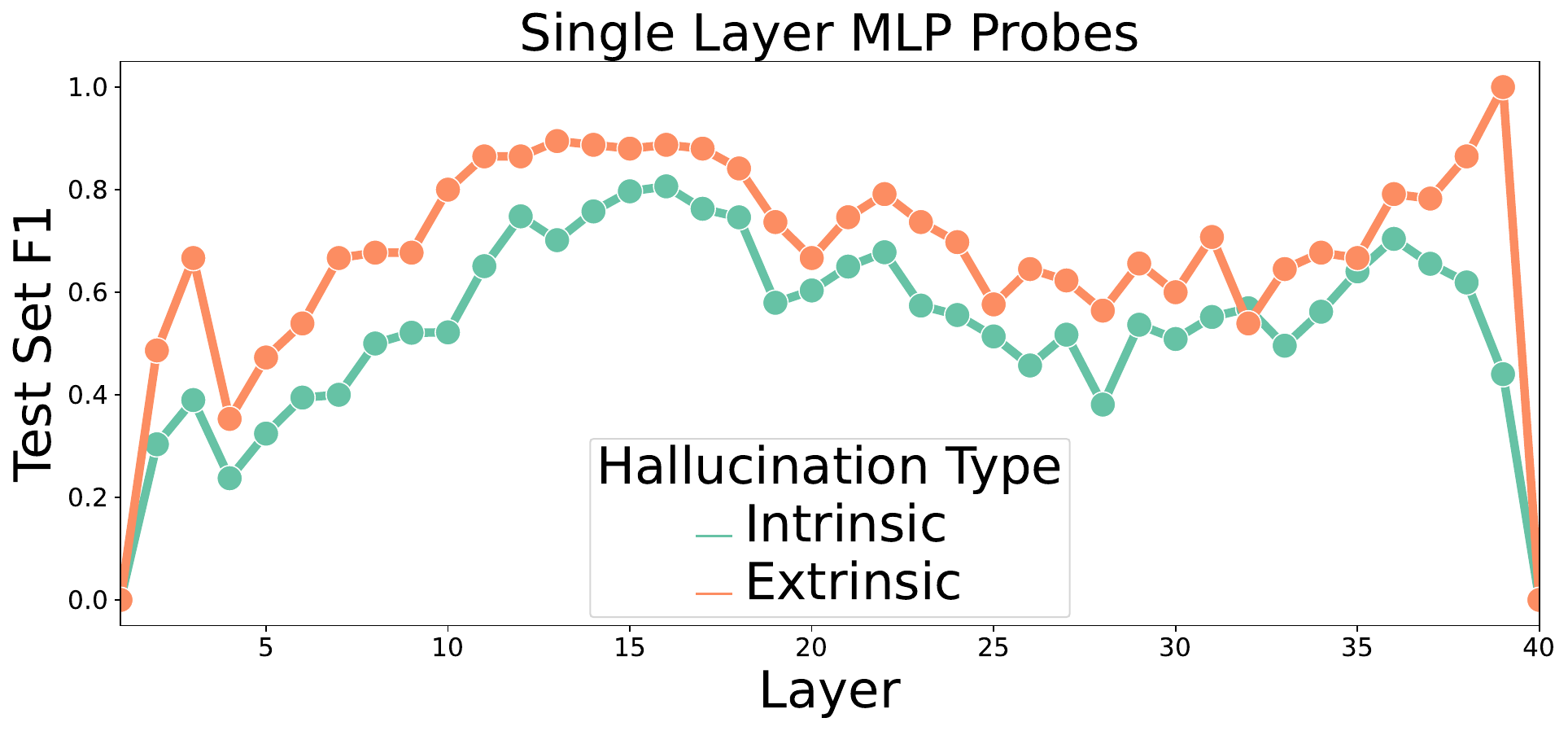} \\
    \caption{\textbf{Intrinsic} and \textbf{extrinsic} hallucination saliency across layers for synthetic CDM as measured by response-level MLP probe F1 (\textbf{F1-R})  on hallucination types; additional plots in \cref{sec:expandedresults},  \cref{fig:type_all}.}
    \label{fig:type}
\end{figure}

\paragraph{Hallucination Type.} Next, we test for the saliency
of different forms of hallucination behavior in model internal states. Here, with annotator-labeled test set responses according to whether they contained intrinsic or extrinsic hallucinations,%
we evaluate trained probe performance
in correctly detecting the presence of hallucinations for instances of each hallucination type.  Intrinsic hallucinations are rarer and may reflect a misunderstanding of the grounding data.  
We observe in \cref{fig:type} that extrinsic hallucinations---which are $1.8\times$ more common at the response level (as labeled by annotators) and presumably reflect a failure to find desired information in the grounding data at all---are easier to detect at every layer. Both kinds are generally most salient in probe middle layers, between layers 10 to 20.
Overall, extrinsic hallucination remained more salient in hidden states than intrinsic hallucination.

\paragraph{Synthetic Hallucinations.}
As \cref{tab:organic_synthetic} shows, probes achieve very high F1 on the detection of synthetically created hallucinations across all tasks. However, probe performance is subject to a drop
when trained and tested on different modalities, with probes generally performing better when trained and tested on the same hallucination modality. 
The most severe version of this is on CDM, which may be caused by differences between our organic annotation instructions and those of the BUMP dataset \cite{ma-etal-2023-bump}.
Prior work has shown that synthetic data generation approaches specifically designed for factuality evaluation \cite{ma-etal-2023-bump} do not align with actual errors made by generation models in their lexical error distributions \cite{goyal-durrett-2021-annotating}. Here, our results show this to extend to their hidden state \textit{signals} as well, which are used by the probe to detect unfaithfulness in the case of forced decoding on human-created content, relative to organic instances of hallucination. 
This is also related to the idea of ecological validity \cite{de2020towards}. Findings from an ecologically valid experiment should hold outside the context of that study. We do not find strong evidence that the detection of synthetic hallucinations generalizes outside that context.
\begin{table}
\centering
\begin{tabular}{ l l l l }
    \toprule
    \textbf{Task} & \textbf{Train} & \textbf{Test} & \textbf{F1-R} \\ 
    \midrule
    CDM         & Organic & Organic & \bf 0.75  \\  
                & Synthetic & Organic & 0.66 \\
                \cline{2-4}
                & Organic & Synthetic & 0.17 \\  
                & Synthetic & Synthetic & \bf 0.86 \\
                \midrule
    CF          & Organic & Organic & \bf 0.94 \\
                & Synthetic & Organic & 0.83 \\
                \cline{2-4}
                & Organic & Synthetic & 0.90 \\
                & Synthetic & Synthetic & \bf 0.95 \\
                \midrule
    E2E         & Organic & Organic & \bf 0.90 \\
                & Synthetic & Organic & 0.71 \\
                \cline{2-4}
                & Organic & Synthetic & 0.94 \\
                & Synthetic & Synthetic & \bf 0.99 \\
    \bottomrule
\end{tabular}
\caption{Response-level Pooling\textsubscript{E} probe performance, as measured by \textbf{F1-R}, in detecting hallucinations on organic or synthetic ungrounded responses.%
}
\label{tab:organic_synthetic}
\end{table}

\paragraph{Lack of Task Transfer.}
To test whether the 
probed features persist in similar hidden state locations across different tasks, we test our probe on each specific task when trained on the others. 
As shown in \cref{tab:crossdomain} in \cref{sec:expandedresults}, %
we find a lack of generalization in hallucination detection performance for our probes across different training and test tasks. 
This may only mean that each probe is overfitting to task-or-dataset-specific confounds.  If so, training on a mixture of many tasks (more than the two we tried in \cref{tab:crossdomain}) might result in a classifier that identifies the consistent hallucination signal and thus generalizes well to new tasks. However, it is also possible that no consistent hallucination signal exists.\looseness=-1

\section{Conclusion}

We examined how an LLM's internal representations reveal its hallucinatory behavior, via a case study of multiple in-context grounded generation tasks.  In specific domains, we can accurately identify hallucinatory behavior through trained probes.  Probe accuracy can depend on model layer, hidden state type, model size, hallucination type, and training-test mismatch. \textit{\footref{q1label}TL;DR: Yes, they do.}

\section*{Limitations}\label{sec:discussion}

Below, we highlight a few takeaways, broader implications, and limitations that naturally suggest directions for future work.

\paragraph{Hallucinatory Span Detection.} In the context of use cases such as retrieval-augmented generation (RAG), span-level results have the potential to be extremely useful because they allow us to detect where hallucinations may fall. For example, hallucinatory spans could be (1) highlighted in the UI for RAG-based applications, (2) detected as soon as they are generated, triggering the decoder to backtrack and regenerate, or (3) a local hallucination detection could be used for further mitigation of hallucination in a reward-shaping setting for RLHF, in contrast to only having response-level signals available. 

\paragraph{Efficiency and Access.} Our work is motivated by the need for efficient LLM hallucination evaluation. Though computationally efficient, one of the key limitations of probing is in the need for labeled in-domain data for probe training.  Our results in \cref{tab:crossdomain} suggest poor generalization to out-of-domain tasks, although perhaps this would be mitigated by larger and more diverse training sets. Furthermore, it perhaps goes without saying that probing requires access to the hidden states themselves. If LLMs continue to move behind closed-source APIs, such hidden state probes will not be possible for those without privileged access.

\paragraph{Annotator Disagreements.} Labeling for faithfulness errors remains a challenging task for annotators \cite{ladhak2024faithfulness}. Despite asking annotators to label what they deem to be the \textit{minimal spans} of hallucination in a response and providing a considerably detailed annotation schema (\cref{sec:annotationguidelines}), sizable variation still exists between annotators on judgments of their exact span boundaries. It remains an open question how to reduce or circumvent annotator disagreement in this context. 

\paragraph{Probe Design.}  Even if span boundaries were standardized, they may not align perfectly with hallucination signals in LLM internal representations. For instance, while a generation may first contradict a given knowledge source at a certain token, the ultimate \textit{signal} that it will hallucinate may already be salient in the tokens preceding it.  Alternatively, the signal might be computed only later, as the decoder encodes previously generated tokens and recognizes that it is currently pursuing an ungrounded train of thought.  Thus, there may be probe architectures that could improve hallucination detection performance and mitigation.  Nonlinear probes such as feed-forward networks are also worth considering: the hidden layer(s) of a feed-forward network could learn to detect different types of hallucinations.
Furthermore, the lack of task transfer that we currently observe presents interesting future challenges in increasing the generalizability of our approach in search of a more consistent signal of hallucination, if one exists.

\paragraph{Mitigation.}
Finally, once we can  \emph{detect} ungrounded generation behavior, we can try to learn to \emph{avoid} it.  This has great practical importance as retrieval-oriented generation becomes popular across a variety of consumer-facing applications. 

We have left this subject for future work.  A simple method is to reject a response or prefix where hallucinations have been detected with high probability, and start sampling the response again \cite{guerreiro-etal-2023-looking}.  Another approach would be to use the predicted hallucination probabilities to reweight or filter hypotheses during beam search.  Finally, a detector can also be used as a proxy reward method for fine-tuning the language model, using a reward-based method such as PPO \cite{schulman2017proximal}.
In addition to learning from response-level negative rewards incurred at the end of the utterance, PPO can benefit from early token-level negative rewards, which serve as a more informative signal for reinforcement learning (i.e., reward shaping) and thus can reduce sample complexity.

 \section*{Acknowledgments}
 We thank Val Ramirez, Patrick Xia, Harsh Jhamtani, Hao Fang, Tongfei Chen, Radhika Gaonkar, and Tatsunori Hashimoto for their helpful comments, thoughts, and discussions, as well as the data specialists who contributed and aided in the creation of this work. 

\newpage

\bibliography{anthology,custom}
\bibliographystyle{acl_natbib}

\clearpage

\appendix

\section*{Appendix}

Here, we provide additional information about annotation guidelines, annotator disagreement examples, prompts used to generate organic responses to our tasks, ChatGPT baseline comparisons, and expanded results.

\section{Annotation Guidelines}
\label{sec:annotationguidelines}

The purpose of this annotation task is two-fold: (1) identify if a generated response is a hallucination relative to a given knowledge source, and (2) if so, identify and mark the location(s) of the minimal spans of text that are ungrounded against the knowledge source. A response is grounded in a knowledge source if that source entails the content of the response. In other words, the response should (1) not contradict or (2) draw conclusions that are not explicitly given in the knowledge source.

\subsection{CNN / DailyMail}

\paragraph{Provided Information.} (1) Original article, (2) summary of article.

\paragraph{Task.} Identify the presence and location of hallucinations in the summary of the article, relative to the original article text. You are to only annotate—by copying the text in the summary and adding characters < and > in specific locations—around the minimal span(s) of the summary text relative to the original article; I.e., if changing an additional token wouldn’t change whether or not the response was grounded or not, don’t include it in the span you annotate. Note: You will be given summaries that contain a variety number of sentences. You are only to annotate the \textit{first three sentences} in the summary, discarding the rest.

\paragraph{Example.} (CNN) -- The world's fastest man on a pair of skis just got faster. On Monday Italian Simone Origone broke his own speed skiing world record as he reached 252.4 kilometers per hour on the Chabrieres slopes in the French Alps -- an achievement confirmed by organizers France Ski de Vitesse. With a 1,220 m slope that has a maximum gradient of 98\% and an average of 52.5\%, Chabrieres is not for the faint-hearted. Traveling at over 250 km/h is a privilege usually reserved for Formula One drivers, yet speed skier Origone was equipped with just an aerodynamic helmet to increase streamlining and a ski suit made from air-tight latex to reduce wind resistance. Origone's one nod to safety was wearing a back protector in case of a crash as he threw himself down the one kilometer track. Origone has been the fastest speed skier on the globe since April 2006, having set a then-new world record of 251.4 km/h at Les Arcs. Bastien Montes of France was Origone's closest challenger on Monday, but even the Frenhcman's new personal best of 248.105 km/h was someway short of the Italian's 2006 world record, let alone his latest one. "Simone Origone is the greatest champion of all time, he is the only person to hold the record for two ski speeds in France -- Les Arcs and Vars Chabrieres. This is a historic day," technical director of Speed Masters Philippe Billy told Vars.com. The 34-year-old Origone -- a ski instructor, mountain guide and rescuer by day -- only took up the discipline in 2003, having given up downhill skiing in 1999. "Now that I have twice won the world record, I can say that I have made history," Origone told Vars.com. "It is important for me and for speed skiing." 

\paragraph{Hypothetical Instance 1.} The world’s fastest man on skis just got faster.
\begin{itemize}
    \item Everything is grounded, no need to mark anything here.
\end{itemize}

\textbf{Annotation.} The world’s fastest man on skis just got faster.

\paragraph{Hypothetical Instance 2.} The world’s fastest man on skis just got faster. He has now won the world record twice. He has excelled at Chabrieres. He has been the fastest speed skier on the globe since April 2006. His speed record reached more than 260 kilometers per hour. 
\begin{itemize}
    \item This summary has more than 3 sentences, so the first thing to do is to only look at the first three and discard any sentences beyond that in your annotation.
    \item Even though the last sentence is ungrounded relative to the source text, it is not within the first three sentences, so we can ignore it here.
    \item Annotate as such, copying the first three sentences into the annotation box. No need to highlight anything, as everything in the first three sentences are grounded:
\end{itemize}

\textbf{Annotation.} The world’s fastest man on skis just got faster. He has now won the world record twice. He has excelled at Chabrieres.

\paragraph{Hypothetical Instance 3.} Simone Origone, \underline{now in his twenties}, just got faster. He has now won the world record \underline{more than twice}. 
\begin{itemize}
    \item ``In his twenties'' contradicts with ``The 34-year-old Origone'' in the original article.
    \item ``more than twice'' contradicts with how he has now only won it exactly twice. As such, annotate:
\end{itemize}

\textbf{Annotation.} Simone Origone, now in his <twenties>, just got faster. He has now won the world record <more than> twice.

\paragraph{Special Points of Note.} Below, we list a few points of special note to keep in mind during your annotations.

\paragraph{Concatenation.} When finding yourself annotating something like ``The city's pollution levels have reached or <exceeded ><``very high''> levels for the last...,'' do concatenate the two labeled spans; if there are no characters between them (for example, <exceeded ><``very high''>), concatenating the annotation bounds into (<exceeded ``very high''>) is ideal.

\paragraph{``Was''/``Is''.} If the article says something like ``Woods insisted he is ready to win the famous Claret...'' and the summary said something like ``said he was ready to win the famous Claret...'', ``was'' here would not be considered to be a hallucination, as the summary is recounting what was said in the article, so the past tense here doesn't contradict what was originally said in the article.

\paragraph{When The Article Has ``Contradicting'' Facts.} Articles may sometimes ``correct'' themselves; i.e. the article may have something where it originally states a claim but then corrects it later in the article. For example, ``The NWS -- last May -- initially forecast 13 to 20 ``named storms,'' including seven to 11 hurricanes. Then in August, it dialed down that initial forecast to predict 13 to 19 named storms, including six to nine hurricanes.'' 

In this case, if the summary only states ``The National Weather Service forecast called for 13 to 20 named storms'' (and doesn't mention it was corrected later to 13 to 19), then this is a hallucination as it is corrected later in the article. 

If the summary states ``The National Weather Service forecast initially called for 13 to 20 named storms'' (and doesn't mention it was corrected later to 13 to 19) then this is not a hallucination, as it quantifies the original prediction with ``initially.''

If the summary states ``The National Weather Service forecast called for 13 to 19 named storms,'' then this is not a hallucination, as it is the final corrected fact. 

\paragraph{Identifying the Summary.} Sometimes the summarization system may produce things that are not a summary of the article. For example, ``Woman dies while paragliding in Tenerife. Paragliding in Tenerife. Paragliding in Tenerife. Paragliding in Tenerife. Paragliding Tenerife Canary Islands.''

Here, only the first sentence is the summary, and you can discard the rest. This is because the summarization system tries to mimic how actual website summaries work, so they might also add things like topic tags, keywords, and things like that.

Another example is ``InvisiBra, £48, Lavalia, lavalia.co.uk; Strapless Stick On Adhesive Bra, £12, Marks and Spencer; Natural Stick On Bra, £20, Debenhams.'' These are all just keyword tags; no summary of the article is present here. Try to use your best judgment here, as it might not always be clear. 

In the case where such things are generated that do not look to be part of a summary of the article in terms of format (not content), \textit{discarding them and \textbf{everything after them} is the way to go.} For the former example, you would annotate only ``Woman dies while paragliding in Tenerife.'' for hallucinations.

\subsection{Conv-FEVER}

\paragraph{Provided Information.} (1) Conversational context, (2) Knowledge for next response, (3) Next response.

\paragraph{Task.} Identify the presence and location of hallucinations in “Next response”, relative to “Knowledge for next response”. You are to only annotate—by copying the text in “Next Reply” and adding characters < and > in specific locations—around the minimal span(s) of an ungrounded response relative to the knowledge; I.e., if changing an additional token wouldn’t change whether or not the response was grounded or not, don’t include it in the span you annotate.

\paragraph{Example.} \textbf{Conversational Context:}
A: lets talk bowling a game where you throw a bowling ball towards a target.
B: ok I love bowling I would like to know more
A: for one it is one of the major forms of throwing sports and I think a lot of people enjoy it
B: yes me too I use to bowl all the time with my family what about the strike?
A: well a strike is when you knock down all pins on the first round
B: ok how can I get a turkey I think that may be hard to get though
\textbf{Knowledge for Next Reply:}
Three consecutive strikes are known as a "turkey".

\paragraph{Hypothetical Instance 1.} \textbf{Next Reply:} A: yea \underline{four} consecutive strikes is known as a turkey
\begin{itemize}
    \item Here, a contradiction is present in the underlined portions: \underline{four} consecutive strikes in the response goes against what is provided in the knowledge, which says \underline{three}. 
    \item To make this reply grounded, one would have to change “four” to “three”. 
    \item As such, in your annotation, you will highlight the minimum needed span of text to make this reply grounded, with chevrons <>. Here, the minimum needed span of text to change is “four”, so you would annotate as such: 
\end{itemize}

\textbf{Annotation.} yea <four> consecutive strikes is known as a turkey

\paragraph{Hypothetical Instance 2.} \textbf{Next Reply:} A: \underline{yea it is pretty hard}, four consecutive strikes is known as a turkey
\begin{itemize}
    \item Here, the contradiction is still present in the previously underlined portion: four is a contradiction with three. 
    \item However, in addition to this contradiction, you see an expression of an opinion in this response: that getting a turkey is pretty hard (underlined). 
    \item You are to ignore the expression of personal opinions during your annotation of minimal spans for this task, as we won't consider conversational chitchat—if complemented with grounded information elsewhere in the reply—to be a hallucination for this task. As such, you would annotate this as:
\end{itemize}

\textbf{Annotation.} yea it is pretty hard, <four> consecutive strikes is known as a turkey

\paragraph{Hypothetical Instance 3.} \textbf{Next Reply:} A: \underline{yea it is pretty hard.}

\begin{itemize}
    \item Here, the response is \underline{purely} an expression of an opinion and does not use any information in the provided knowledge. 
    \item As such, you would mark this entire response to be a hallucination, annotating as:
\end{itemize}

\textbf{Annotation.} <yea it is pretty hard.>

\paragraph{Hypothetical Instance 4.} \textbf{Next Reply:} A: Three strikes consecutively is a turkey.

\begin{itemize}
    \item Here, the response is grounded. As such, you won’t mark anything in the response, annotating as such:
\end{itemize}

\textbf{Annotation.} Three strikes consecutively is a turkey.

\paragraph{Hypothetical Instance 5.} \textbf{Next Reply:} A: \underline{four} consecutive strikes is known as a turkey, \underline{also known as a strike.}

\begin{itemize}
    \item Two hallucinations are present in this response:
    \item The direct contradiction of four with three in the knowledge source, and 
    \item The final part of the reply, saying that it’s also known as a strike, which is information not provided anywhere in the knowledge source. As such, you would mark this as:
\end{itemize}

\textbf{Annotation.} <four> consecutive strikes is known as a turkey, <also known as a strike>

\subsection{E2E}

\paragraph{Provided Information.} (1) Restaurant attributes, (2) Generated restaurant description.

\paragraph{Task.} Identify the presence and location of hallucinations in the generated restaurant description—by copying the text in the summary and adding characters < and > in specific locations—around the minimal span(s) of the generated restaurant description relative to the provided attributes. i.e., if changing an additional token wouldn’t change whether the description was grounded or not, don’t include it in the span you annotate. Note that numerical ratings should be rendered as numerical ratings and text ratings should be rendered as non-numerical expressions. For example, “customer rating” can be “low”, “medium” and “high”, but also “1”, “3”, and “5”. If the customer rating is “1”, for instance, and the generated description says “…has a low customer rating”, that would be considered a hallucination. The “priceRange” rating also has this attribute, in addition to “customer rating.” Finally, E2E specifies the "familyFriendly" attribute as akin to Child-friendly in gold annotations: if a restaurant is family-friendly, then it is also child-friendly.

\paragraph{Hypothetical Instance 1.} \textbf{Restaurant Attributes:} name[The Wrestlers], priceRange[less than £20], customer rating[low], familyFriendly[no].
\textbf{Restaurant Description:} The Wrestlers is an adults only restaurant with a customer rating of 1 out of 5 and a price range of less than £20.

\begin{itemize}
    \item While the familyFriendly[no] attribute says that the restaurant is not family friendly, this does not mean that the restaurant is adults only.
    \item -	Having a low customer rating does not imply a 1 out of 5 customer rating. As such, annotate as:
\end{itemize}

\textbf{Annotation.} The Wrestlers is an <adults only> restaurant with a customer rating of <1 out of 5> and a price range of less than £20.

\paragraph{Hypothetical Instance 2.} \textbf{Restaurant Attributes:} name[The Dumpling Tree], eatType[restaurant], food[Italian], priceRange[high]. 
\textbf{Restaurant Description:} The Dumpling Tree is an Italian restaurant with high prices.

\begin{itemize}
    \item Everything is grounded, no need to mark anything. 
\end{itemize}

\textbf{Annotation.} The Dumpling Tree is an Italian restaurant with high prices.

\paragraph{Hypothetical Instance 3.} \textbf{Restaurant Attributes:} name[The Wrestlers], priceRange[less than £20], customer rating[low], familyFriendly[no].
\textbf{Restaurant Description:} Prices will never exceed £20 at The Wrestlers and they also don't really rate well. They are not even family-friendly.

\begin{itemize}
    \item Having a priceRange attribute of less than 20 does not mean that prices will never exceed twenty.
    \item As “will never exceed” is the minimal span you would need to change to make the description grounded relative to the provided attributes, annotate:
\end{itemize}

\textbf{Annotation.} Prices <will never exceed> £20 at The Wrestlers and they also don't really rate well. They are not even family-friendly.

\paragraph{Hypothetical Instance 4.} \textbf{Restaurant Attributes:} name[Midsummer House], food[French], customer rating[5 out of 5], near[Café Rouge].
\textbf{Restaurant Description:} A place for fine dining French food, Midsummer House is nearby Cafe Rouge.

\begin{itemize}
    \item Nowhere in the attributes does it say that it’s “fine dining.” As such, annotate:
\end{itemize}

\textbf{Annotation.} A place for <fine dining> French food, Midsummer House is nearby Cafe Rouge.

\paragraph{Hypothetical Instance 5.} \textbf{Restaurant Attributes:} name[Browns Cambridge], priceRange[cheap], priceRange[moderate], customer rating[1 out of 5].
\textbf{Restaurant Description:} The Brown's chain of food restaurants is an 1 out of 5 customer rated restaurant, and have moderate and cheap pricing. They are childfree.

\begin{itemize}
    \item The attribute does not mention that Browns Cambridge is a chain of food restaurants. 
    \item Furthermore, it does not say that the restaurant is childfree. As such, annotate:
\end{itemize}

\textbf{Annotation.} The Brown's <chain of food restaurants> is an 1 out of 5 customer rated restaurant, and have moderate and cheap pricing. <They are childfree.>

\paragraph{Hypothetical Instance 6.} \textbf{Restaurant Attributes:} name[Browns Cambridge], priceRange[more than £30].
\textbf{Restaurant Description:} The Browns Cambridge is a restaurant with high prices.

\begin{itemize}
    \item More than £30 does not necessarily mean “high prices.” As such, annotate:
\end{itemize}

\textbf{Annotation.} The Browns Cambridge is a restaurant with <high prices>.

\begin{table}
\centering
\small
\begin{tabular}{p{0.9\linewidth}}
\toprule \\ 
 Taxi driver dropped off a soldier on an unsafe road, and he was killed by a coach. \textbf{The taxi driver} was not at fault.
 \\ \\
 Taxi driver dropped off a soldier on an unsafe road, and he was killed by a coach. \textbf{The taxi driver was not at fault.}
 \\\\ \midrule \\
 
 Accounting is the \textbf{language of business, and it measures the results of an organization's economic activities.}
 \\ \\ 
 \textbf{Accounting is the language of business, and it measures the results of an organization's economic activities.}
 \\\\  \midrule \\

 Alimentum is a \textbf{high rated} family-unfriendly restaurant with less than £20 per person.
 \\ \\ 
 Alimentum is a \textbf{high} rated family-unfriendly restaurant with less than £20 per person.
 \\ \\ \bottomrule
\end{tabular}
\caption{Annotator span-level annotation differences for hallucination, annotated in \textbf{bold}, in organic model responses for CNN / DailyMail (top), Conv-FEVER (middle), and E2E (bottom).}
\label{table:spandiffs}
\end{table}

\section{Annotator Disagreements}
\label{sec:annotatordisagreements}

Examples of annotator disagreements in the exact spans of hallucination are shown in \cref{table:spandiffs}. Qualitatively, we find that annotators agreed \textit{generally} on the specific items of hallucination; i.e. the subject that was hallucinated (i.e. the taxi driver), the description of what was mentioned (i.e. the description of accounting), and the attribute that was hallucinated (i.e. the specific rating of the resturant). However, \textit{where} exactly those hallucinated spans began and ended was a source of reasonable variation among annotators. 

\section{Prompts and Further Evaluation Details}
\label{sec:prompts}

\paragraph{Organic Model Responses.}
CNN / DailyMail generations take the form of a 0-shot prompt with \texttt{[ARTICLE] TL;DR: [SUMMARY]}, whereafter the first 3 sentences generated in the summary are taken to be the summary of the article. For Conv-FEVER, we use a 2-shot in-context learning prompt in the form \smallskip

\begin{displayquote}
\raggedright

\texttt{\#\# CONVERSATION HISTORY }

\texttt{[CONTEXT]}

\texttt{\#\# KNOWLEDGE FOR NEXT RESPONSE}

\texttt{[KNOWLEDGE]}

\texttt{\#\# RESPONSE THAT USES THE INFORMATION IN KNOWLEDGE}

\texttt{[RESPONSE]} \smallskip

\end{displayquote}

\noindent whereafter everything generated in the response preceding a newline character is taken as the final generation. For E2E, we use a 10-shot in-context learning prompt in the form of \smallskip

\begin{displayquote}
\raggedright

\texttt{Restaurant Attributes: [RESTAURANT ATTRIBUTES]}

\texttt{Restaurant Description: [RESTAURANT DESCRIPTION]} \smallskip

\end{displayquote}

with the same newline constraints as Conv-FEVER.

\paragraph{FActScore Evaluation.} As FActScore \cite{factscore} was originally designed for evaluating the faithfulness of long-form generated texts, we slightly modify its evaluation for our Data-to-Text task, E2E. Specifically, instead of evaluating on raw attribute sets (i.e. \texttt{name[Clowns], eatType[pub], near[The Sorrento]}), we transform these attributes into natural language sentences via simple regular expressions. In this example, these attributes would instead be transformed into:

\begin{displayquote}
The name of this restaurant is Clowns. This restaurant is a pub. This restaurant is near The Sorrento.
\end{displayquote}

For our other tasks on abstractive summarization and knowledge-grounded dialogue generation, we keep evaluations unchanged with the original implementation of FActScore for our evaluations.

\subsection{ChatGPT Prompts}
\label{sec:chatgptprompts}

\paragraph{ChatGPT Classic.} For the most basic zero-shot prompt classifier with no reasoning, we use the following prompt for \textbf{CNN / DailyMail}, making minimal changes relative to \citet{luo2023chatgpt}: \smallskip

\begin{displayquote}
\raggedright

\texttt{Decide if the following summary (SUMMARY) is consistent with the corresponding article (ARTICLE). Note that consistency means all information in the summary (SUMMARY) is supported by the article (ARTICLE).}

\texttt{Article (ARTICLE): [Article]}

\texttt{Summary (SUMMARY): [Summary]}

\texttt{Answer (yes or no):} \medskip

\end{displayquote}

For \textbf{Conv-FEVER}, our prompt is structured as follows: \smallskip

\begin{displayquote}
\raggedright

\texttt{Decide if the following sentence (SENTENCE) is consistent with the corresponding knowledge statement (KNOWLEDGE). Note that consistency means all information in the sentence (SENTENCE) is supported by the knowledge statement (KNOWLEDGE).}

\texttt{Knowledge Statement (KNOWLEDGE): [Knowledge]}

\texttt{Sentence (SENTENCE): [Response]}

\texttt{Answer (yes or no):} \medskip

\end{displayquote}

Finally, for \textbf{E2E}, our prompt is: \smallskip

\begin{displayquote}
\raggedright

\texttt{Decide if the following restaurant description (DESCRIPTION) is consistent with the provided restaurant attributes (ATTRIBUTES). Note that consistency means all information in the description is supported by the provided attributes.}

\texttt{Restaurant Attributes (ATTRIBUTES): [Attributes]}

\texttt{Restaurant Description (DESCRIPTION): [Response]}

\texttt{Answer (yes or no):}

\end{displayquote}

\paragraph{ChatGPT CoT.} For zero-shot chain-of-thought reasoning prompts, we simply replace \texttt{Answer (yes or no):} in all prompts with \texttt{Explain your reasoning step by step then answer (yes or no) the question:}, following \cite{luo2023chatgpt}.

\section{Expanded Results and Discussions}
\label{sec:expandedresults}

\begin{figure}[t]
    \centering
    \includegraphics[width=0.32\textwidth]{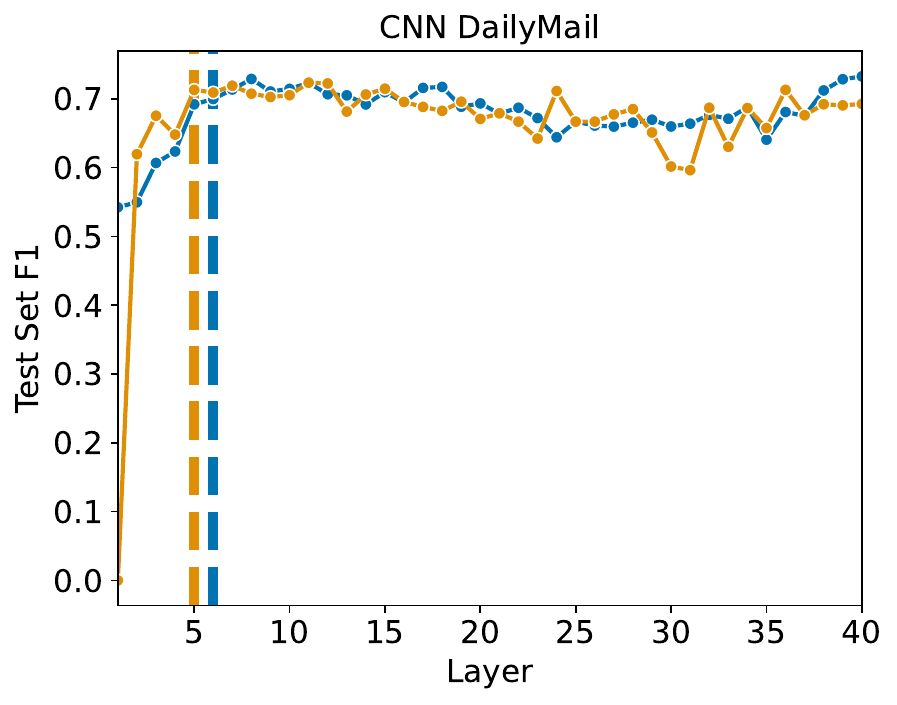}
    \includegraphics[width=0.32\textwidth]{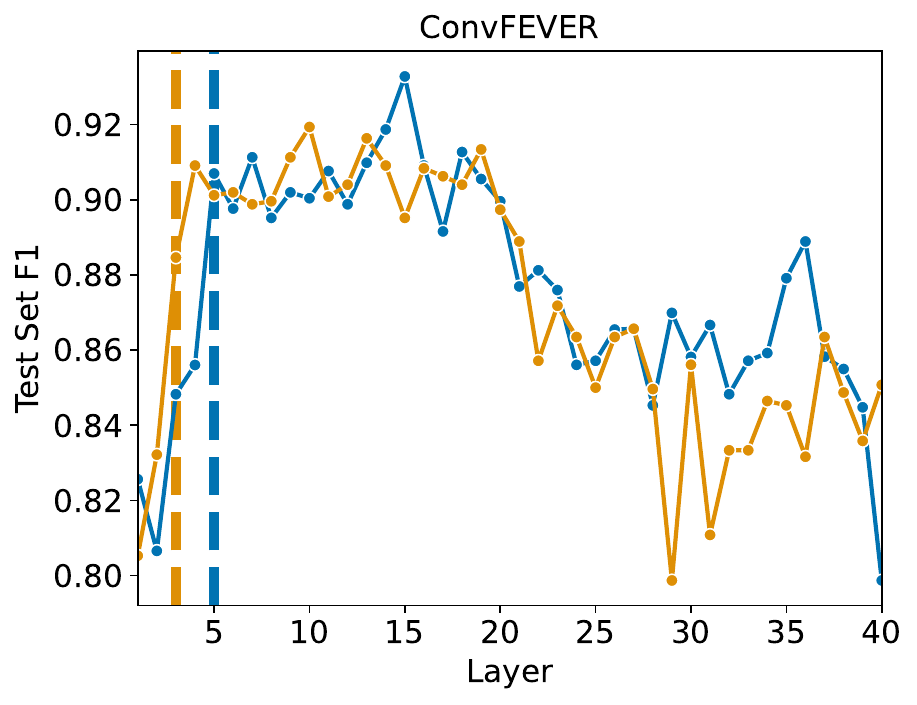}
    \includegraphics[width=0.32\textwidth]{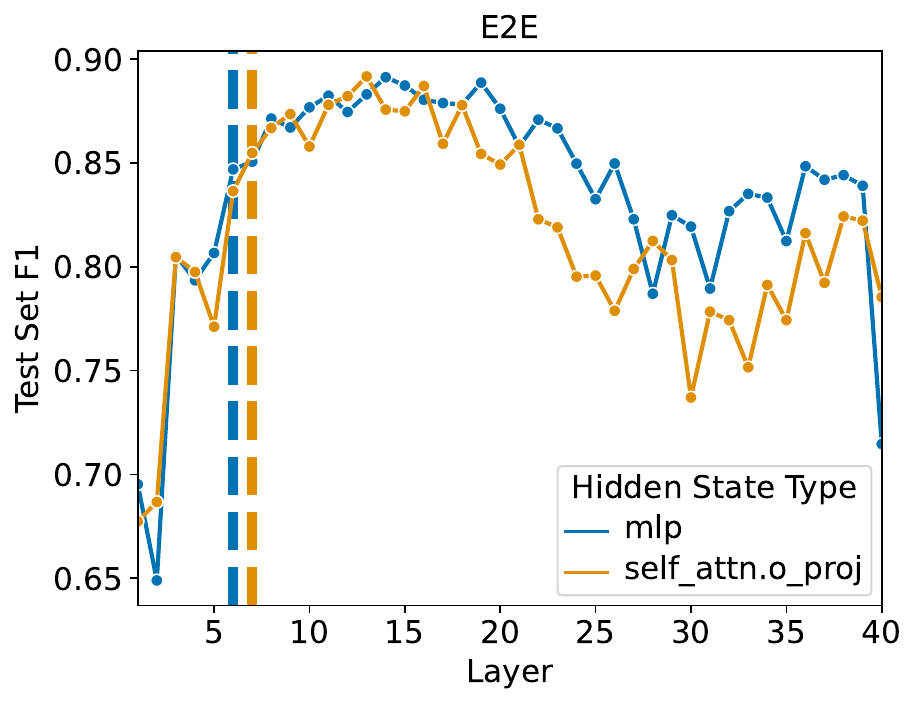}
    \caption{Grounding behavior saliency—as measured by probe response-level F1 on test set—of single layer probe classifiers trained on the \textbf{attention-head out projections} and \textbf{feed-forward states} of \texttt{llama-2-13b}, stratified across tasks and layers. Vertical dashed lines denote the layers that respective probes first surpass 95\% of their peak performance.
    }
    \label{fig:layers_types_all}
\end{figure}

\begin{figure}[t]
    \centering
    \includegraphics[width=0.40\textwidth]{figures/intrinsic_extrinsic_mlp_f1.pdf} \\
    \includegraphics[width=0.40\textwidth]{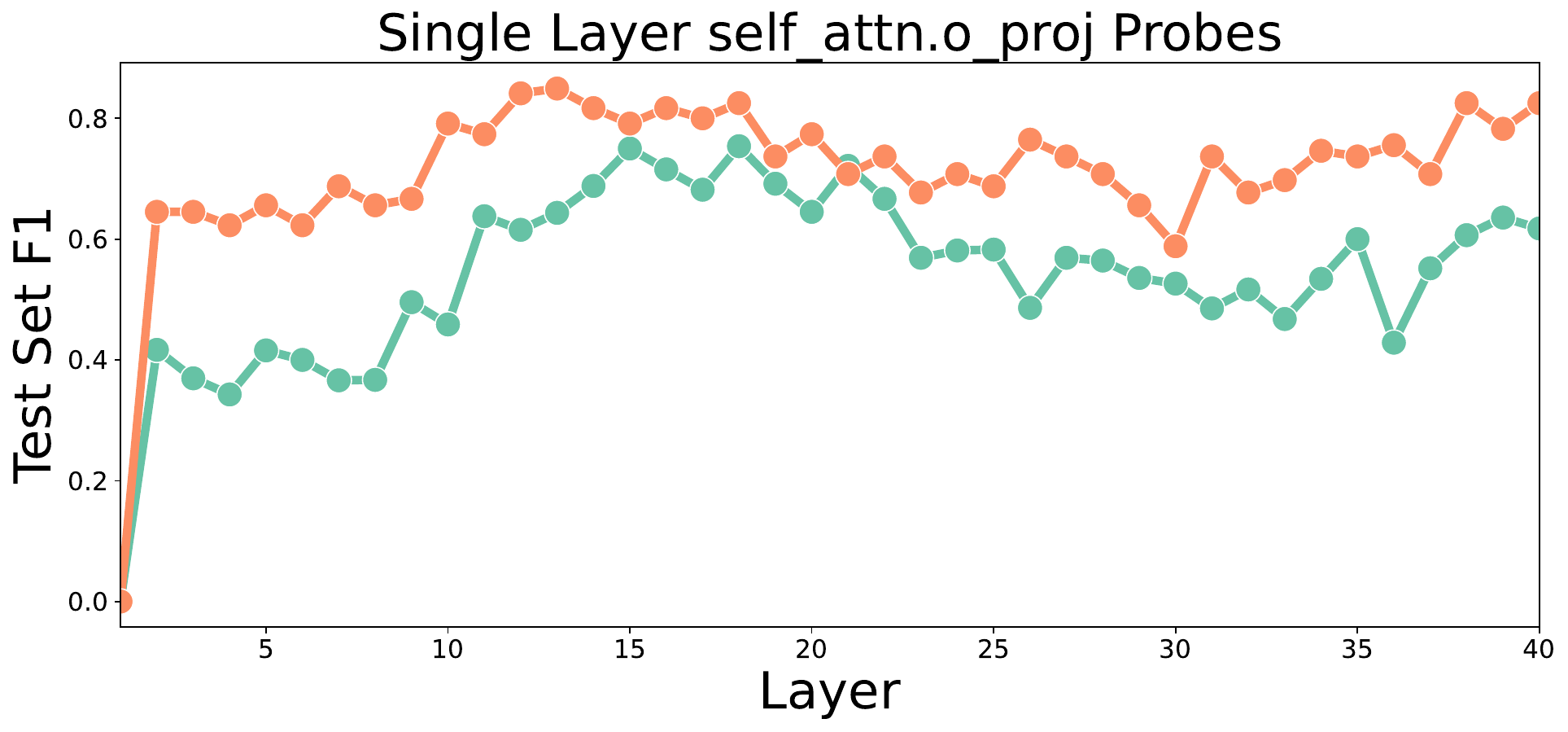}
    \caption{\textbf{Intrinsic} and \textbf{extrinsic} hallucination saliency across layers for synthetic abstractive summarization (BUMP, for CDM) as measured by response-level MLP and Attention probe F1 on hallucination types.}
    \label{fig:type_all}
\end{figure}

\paragraph{Grounding Behavior Saliency.} Full saliency results for all tasks and layers are shown in \cref{fig:layers_types_all}; saliency as stratified across hallucination layers and types across both MLP and Attention probes is shown in \cref{fig:type_all}.

\paragraph{Task Transfer.} Full results for task transfer using the response-level Pooling\textsubscript{E} probe are shown in \cref{tab:crossdomain}.

\begin{table}
\centering
\begin{tabular}{ l l l }
    \toprule
    \textbf{Test} & \textbf{Train} & \textbf{F1-R} \\ 
    \midrule
    CDM & CF & 0.67 \\
        & E2E & 0.65 \\
        & CF+E2E & 0.68 \\
        \cline{2-3}
        & CDM & \bf 0.70 \\
    \midrule
    CF  & CDM & 0.84 \\
        & E2E & 0.81 \\
        & CDM+E2E & 0.84 \\
        \cline{2-3}
        & CF & \bf 0.93 \\
    \midrule
    E2E & CDM & 0.06 \\
        & CF & 0.71 \\
        & CDM+CF & 0.71 \\
        \cline{2-3}
        & E2E & \bf 0.86 \\
    \bottomrule
\end{tabular}
\caption{Response-level Pooling\textsubscript{E} cross-task probe performance, as measured by \textbf{F1-R}. All experiments here use the same amount of training data (a random subset of the full training set).
The "+" sign indicates an equal mixture of two training tasks.}
\label{tab:crossdomain}
\end{table}

\paragraph{Differences from Specific Prior Work.} In the review process, reviewers raised uncertainties about the specific differences between our work and that of \citet{azaria2023internal} and \citet{kadavath2022language}; we note our response here.

\citet{azaria2023internal} are interested in detecting the truth or falsehood of factual statements (e.g., ``H2O is water, which is essential for humans''). The models they probe are presented with these statements without any grounding knowledge, so the probe can only measure if the model’s hidden states reveal some discrepancy between the encoded statement and the training data as encoded by the learned weights.  This cannot work unless the learned weights really encode the truth, so one should only test facts that are well-known and unchanging or slowly changing in time, modality, etc. 

In contrast, a RAG chatbot built on a company’s internal data will be discussing 
specialized and often ephemeral retrieved information that is not well represented in the world knowledge of an LLMs weights. Here, it is important to ensure that the model appropriately follows task instructions, and 
in particular that it does not hallucinate in the sense defined by \cref{sec:intro} and \citet{ji2023survey}: its responses should be appropriately grounded in the retrieved data. 

For emphasis we highlight three important distinctions. (1) \citet{azaria2023internal} examine the veracity of statements without supporting text, while we examine hallucination with respect to grounding text and task instructions. (2) They do not probe models on text that the models themselves generate, but text provided by a secondary model (ChatGPT) and additional human curation. This is an important distinction—we evaluate hallucination on both organically generated texts (and their hidden states) and those generated by a secondary model + human manual curation (synthetic grounding errors) and find that the probe performance generally degrades when probing across these two cases. Finally, (3) they do not study span or token-level detection, which is one of our primary contributions.

\citet{kadavath2022language} similarly probe an LLM in a closed book QA task by training an attention head to detect when the model does not know the answer to a trivia question in the TriviaQA dataset \cite{joshi-etal-2017-triviaqa}. This is different than detecting hallucination in a grounded generation task, as it only tests recall of knowledge already stored in model weights. As for \citet{azaria2023internal}, their generation task does not require the model to draw information from evidence in the prompt that may disagree with stored knowledge.

\end{document}